%% file: main.tex
\documentclass[letterpaper, 10 pt, conference]{ieeeconf}  

\IEEEoverridecommandlockouts                              

\usepackage{epsfig}
\usepackage{graphicx}
\usepackage{color}
\usepackage{hyperref}

\definecolor{turquoise}{cmyk}{0.65,0,0.1,0.1}
\definecolor{purple}{rgb}{0.65,0,0.65}
\definecolor{darkgreen}{rgb}{0.0, 0.5, 0.0}
\definecolor{darkred}{rgb}{0.5, 0.0, 0.0}
\definecolor{darkblue}{rgb}{0.0, 0.0, 0.5}
\definecolor{blue}{rgb}{0.0, 0.0, 1.0}
\definecolor{orange}{rgb}{1.0, 0.5, 0.0}
\definecolor{red}{rgb}{1.0, 0.0, 0.0}

\definecolor{green}{RGB}{3,112,15}

\title{\LARGE \bf
A Protocol for Validating Social Navigation Policies
}

\author{S\"oren Pirk$^{1}$ Edward Lee$^{1}$ Xuesu Xiao$^{2,3}$ Leila Takayama$^{1,4}$  Anthony Francis$^{1}$ Alexander Toshev$^{5}$
\thanks{$^{1}$Robotics at Google}
\thanks{$^{2}$Everyday Robots, Alphabet, \url{www.everydayrobots.com}}%
\thanks{$^{3}$The University of Texas at Austin}%
\thanks{$^{4}$Hoku Labs}%
\thanks{$^{5}$Apple ML Research; work done at Robotics at Google}%
}

\begin{document}

\maketitle
\thispagestyle{empty}
\pagestyle{empty}

\input{./sources/00_abstract.tex}

\input{./sources/01_introduction.tex}
\input{./sources/02_related_work.tex}
\input{./sources/03_method.tex}

\input{./sources/05_experiments.tex}
\input{./sources/06_conclusion.tex}

{\small
\bibliographystyle{ieee}
\bibliography{main}
}

\end{document}

%% file: sources/00_abstract.tex
\begin{abstract}
Enabling socially acceptable behavior for situated agents is a major goal of recent robotics research. Robots should not only operate safely around humans, but also abide by complex social norms. A key challenge for developing socially-compliant policies is measuring the quality of their behavior. Social behavior is enormously complex, making it difficult to create reliable metrics to gauge the performance of algorithms. In this paper, we propose a protocol for social navigation benchmarking that defines a set of canonical social navigation scenarios and an in-situ metric for evaluating performance on these scenarios using questionnaires.
Our experiments show this protocol is realistic, scalable, and repeatable across runs and physical spaces. Our protocol can be replicated verbatim or it can be used to define a social navigation benchmark for novel scenarios. Our goal is to introduce a protocol for benchmarking social scenarios that is homogeneous and comparable. 
\end{abstract}

%% file: sources/01_introduction.tex
\section{Introduction}
One of the main prerequisite of making robots ubiquitous and generally applicable is to endow them with the ability to move around people in a socially acceptable manner. A robot must be able to accomplish navigation tasks while adhering to social norms in shared spaces and respecting human actions and behaviors. We refer to such type of navigation as Social Navigation. Recently, the robotics community has witnessed an increased interest in Social Navigation. Among many other directions, researchers investigate the importance of respecting personal space~\cite{1308100}, maintaining social dynamics~\cite{8036225} and  velocities~\cite{10.1145/2696454.2696463},  socially-acceptable approaching behavior~\cite{huang2014}, and navigation in the presence of groups of people \cite{10.1007/978-3-030-50334-5_24}.   

While these approaches are a testament for the rapid progress in this direction, the evaluation and comparison of algorithms has proven to be difficult for Social Navigation research. To facilitate progress in the community it is of paramount importance to have shared, realistic, repeatable, and scalable benchmarks. If one is to define such a benchmark, then simulation is one tool of choice -- we have seen an increase in robotic simulation environments that  focus on physics and visual realism~\cite{perille2020benchmarking,tsoi2020sean,manso2020socnav1}; simulation is scalable and repeatable across labs. However, simulating the intricacies of human behavior at different  levels of abstraction, ranging from atomic actions and motion dynamics to more complex activities and behavior, has proven to be difficult. Therefore, simulation of humans falls short of providing a realistic medium for social navigation benchmarking.

A different approach for evaluation is to perform demonstrations and studies in uncontrolled real settings in the wild, e.~g.~accessible public spaces such as university campuses. While such evaluations are by definition realistic, they face several limitations. For one, they are not guaranteed to be behaviorally natural as one has to enact social interactions in real studies, which may lead to undesired patterns in the observations (e.g. such as teetering motions). More importantly, though, real experiments are difficult to repeat and to conduct at scale. Every run, even under controlled conditions, will differ from prior runs and running experiments repeatedly -- with human in the loop -- can be extremely costly. Finally, social interactions in real environments are defined by a wide range of variables (e.g. differences in human behavior and appearance, environmental settings, etc.) that make obtaining meaningful measurements difficult. 

\begin{figure}[t]   
  \begin{center}
  \includegraphics[width=1.0\linewidth]{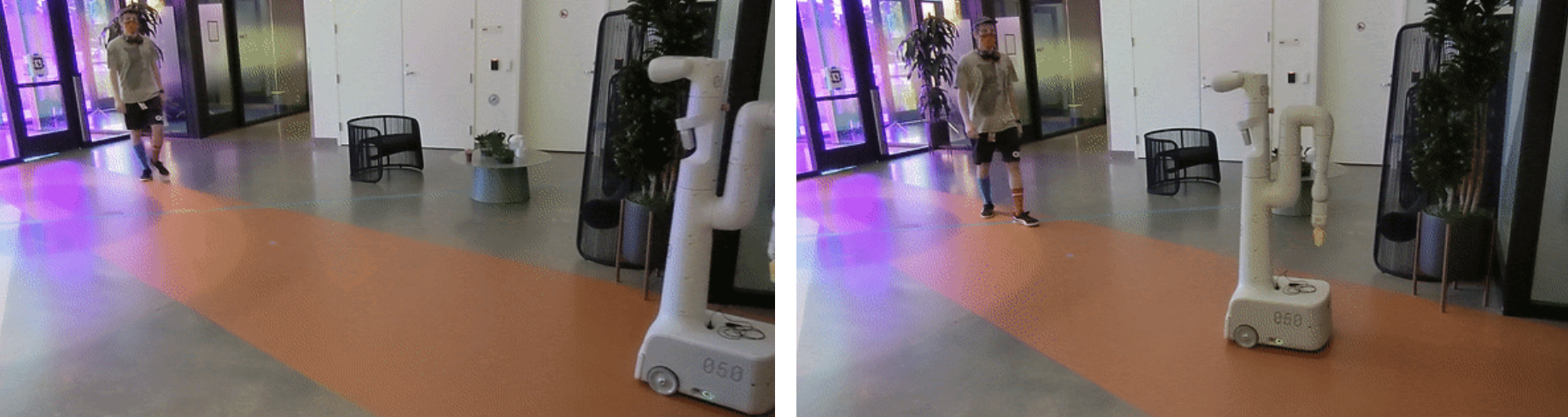} 
  \end{center}
  \vspace{-3mm}
  \caption{\small Frontal approach scenario: human and robot interact by moving along a straight trajectory in opposite directions (left). The robot yields early on to not block the human from walking along their path (right).}
  \label{fig:teaser}   
  \vspace{-6mm}
\end{figure}

\begin{figure*}[ht]   
  \begin{center}
  \includegraphics[width=0.8\linewidth]{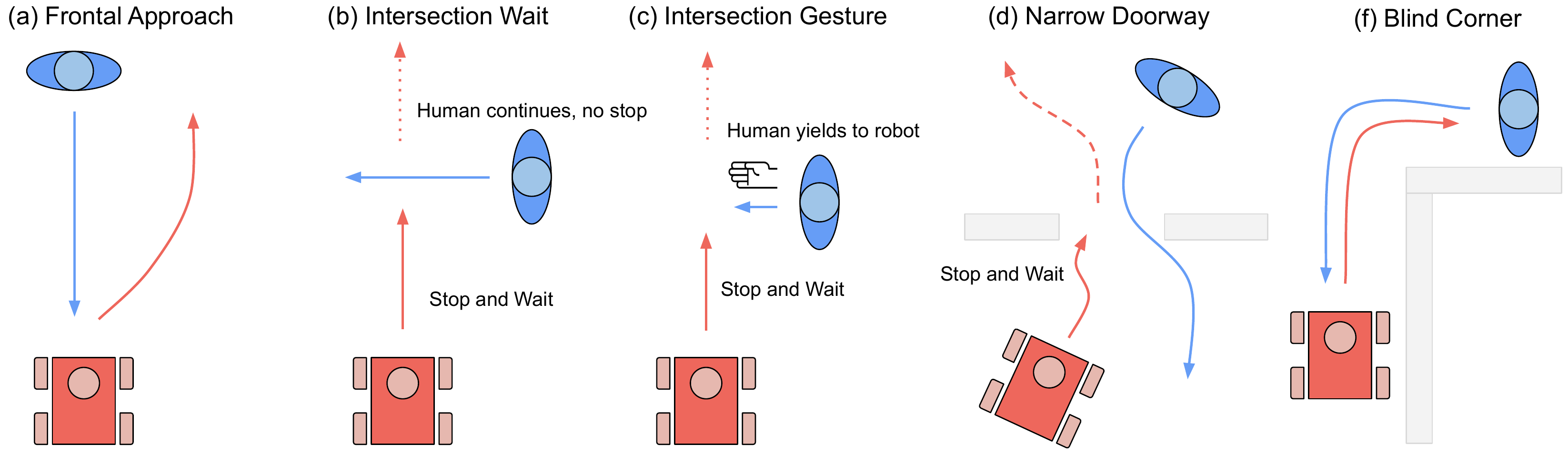} 
  \end{center}
  \vspace{-5mm}
  \caption{The five social navigation scenarios of our benchmark: frontal approach (a), intersection wait (b), intersection gesture (c), narrow doorway (d), and blind corner (e).}
  \label{fig:social_scenarios}   
  \vspace{-6mm}
\end{figure*}

To address these challenges, we aim to propose a protocol for establishing a social navigation benchmark. The desired properties of our benchmark are:
\begin{itemize}
    \item \textbf{Realism}: The benchmark is implemented in a real environment, real robot, and real humans;
    \item \textbf{Scalability}: The benchmark allows for testing on a diverse set of social situations, with a cost which allows for frequent evaluations;
    \item \textbf{Repeatability}: The benchmark is repeatable across different runs and instantiations in different physical spaces.
\end{itemize}

To achieve the above properties we propose a real benchmark based on predefined set of social scenarios evaluated using user surveys. In more detail, we introduce a set of social navigation scenarios (Figure~\ref{fig:social_scenarios}) implemented in real-world settings (Figure~\ref{fig:teaser}). Each scenario is a canonical example of a common human-robot interaction that can occur when performing navigation tasks. The idea is that reducing a human-robot interaction to its essence allows us to better understand and validate social behavior of humans and robots. 
To this end, we define the scenarios in a way that they can be replicated by different labs with low effort and so as to avoid high variance in the human-robot interaction. This addresses the challenge of repeatability as enacting the canonical scenarios will be comparable. 

Second, we propose an in-situ metric based on questionnaires to obtain ratings of humans who experienced the interaction with the robot. Unlike other metrics, this allows us to validate the performance of navigation policies w.r.t. social expectations. We show initial results of validating different policies and that questionnaires can be used to measure meaningful gradients for validating social navigation policies.  
Third, we present guidelines and good practices for defining social navigation benchmarks for other scenarios and environments. Altogether, we hope that our protocol will prove useful for the community to converge to more standardized validation setups.

%% file: sources/02_related_work.tex
\section{Related Work}

Driven by its importance to robotics, social navigation has become the focus of a growing body of research. 
Because of social navigation's complexity and growth, we cannot comprehensively discuss all related work.
For an overview touching on the validation of social navigation, the interested reader is referred to
the recent survey papers of Gao and Huang~\cite{10.3389/frobt.2021.721317}, Charalampous et al.~\cite{CHARALAMPOUS201785}, Mavrogiannis et al.~\cite{mavrogiannis2021core}, Kruse et al.~\cite{KRUSE20131726}, Rios-Martinez~\cite{Rios-Martinez2015}, and more recently, Xiao et al.~\cite{xiao2022motion} and Mirsky et al.~\cite{mirsky2021prevention}. 
Closest to our work are methods that use metrics to measure human discomfort~\cite{6224934,kothari2021,doi:10.1098/rspb.2009.0405} or sociability~\cite{Pacchierotti2005,10.1145/3434074.3447225,10.1145/3415139}, datasets for social navigation~\cite{https://doi.org/10.48550/arxiv.2203.15041}, as well as approaches that employ questionnaires for validation purposes~\cite{Bartneck2009,10.1145/3385121,VEGA201972}.

%% file: sources/03_method.tex
\section{Benchmark for Social Navigation}

\begin{figure}[t]   
  \begin{center}
  \includegraphics[width=1.0\linewidth]{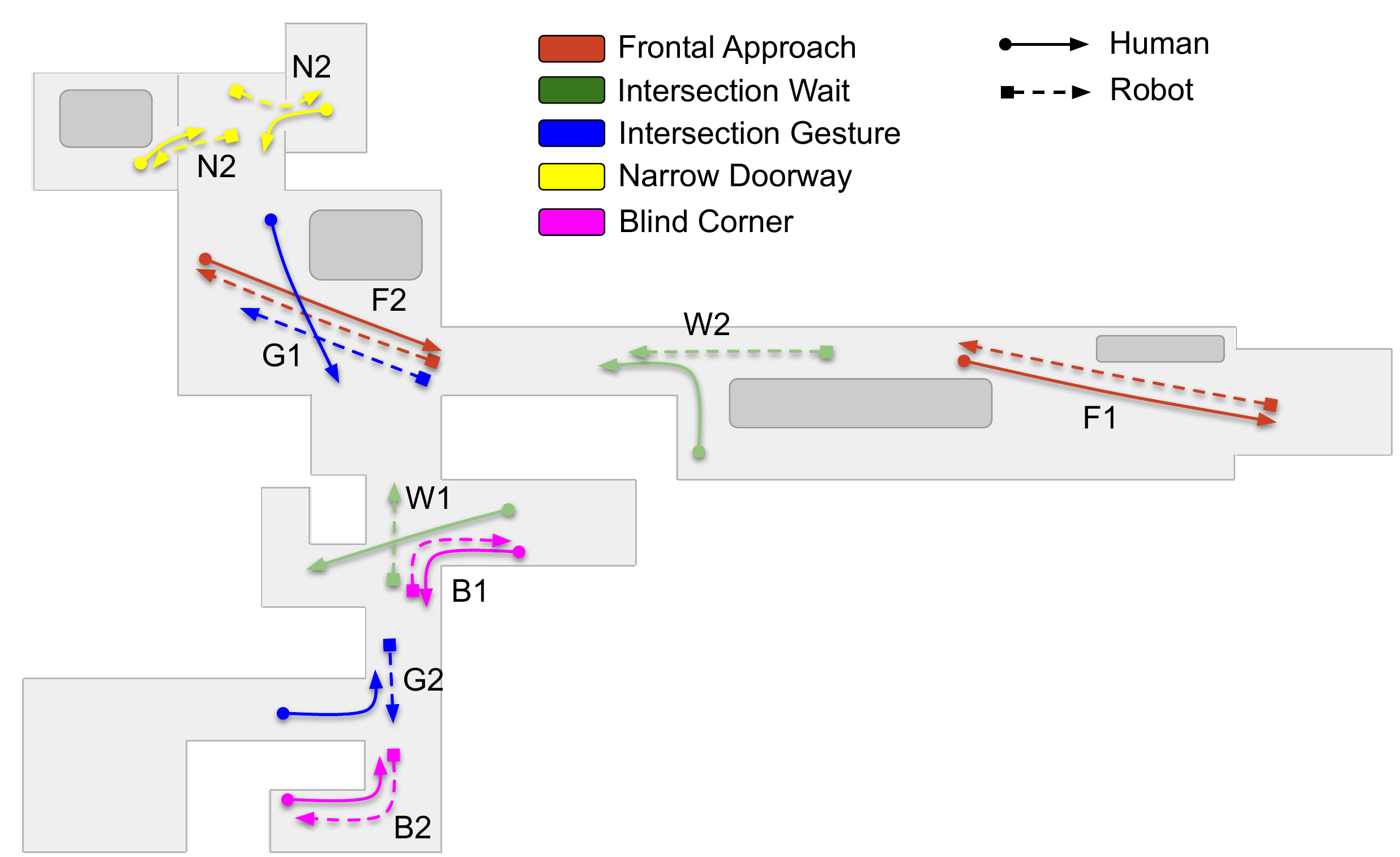} 
  \end{center}
  \vspace{-5mm}
  \caption{Illustration of the locations we used for each of the social scenarios: frontal approach (red), intersection wait (green), intersection gesture (blue), narrow doorway (yellow), and blind corner (magenta). Grey round boxes represent obstacles present in the building (e.g., such as chairs or desks). Arrows with round ends and solid lines indicate human trajectories and arrows with square ends and dashed lines robot trajectories.}
  \label{fig:floor_layout}   
  \vspace{-5mm}
\end{figure}

To establish a social navigation benchmark our goal is to define a set of canonical social navigation scenarios. Each scenario represents a common interaction between a human and a robot performing a navigation task. Specifically, we define \textit{frontal approach}, \textit{intersection wait}, \textit{intersection gesture}, \textit{narrow doorway}, and \textit{blind corner}. For each scenario we define start and end points and task the robot to navigate along the trajectory between the two points (see Fig.~\ref{fig:floor_layout}). Each scenario is then enacted by a human, who is provided with a short description of what is expected to happen. For example, for our frontal approach scenario, we simply say ``Please walk along this trajectory, start walking when the robot is here, the robot is expected to yield.'' The human is then walking in the opposite direction of the robot, while the robot is driving toward its goal position. By defining a specific scenario with a constrained trajectory we define a canonical example of a social interaction.  

Please note that we aim to keep the definition of a social scenario as lightweight as possible -- we only define the social scenario along with a brief description and the start and end points of the trajectory for the robot. We do not specify any constraints for the environment, human behavior, and human appearance. This allows us to implement and then validate the social scenarios in various different environments, while we can also repeatedly measure the performance of a policy on a defined social scenario. To establish a benchmark we define each scenario two times in different locations of our building (Fig.~\ref{fig:floor_layout}).  

\begin{figure}[t]   
  \begin{center}
  \includegraphics[width=1.0\linewidth]{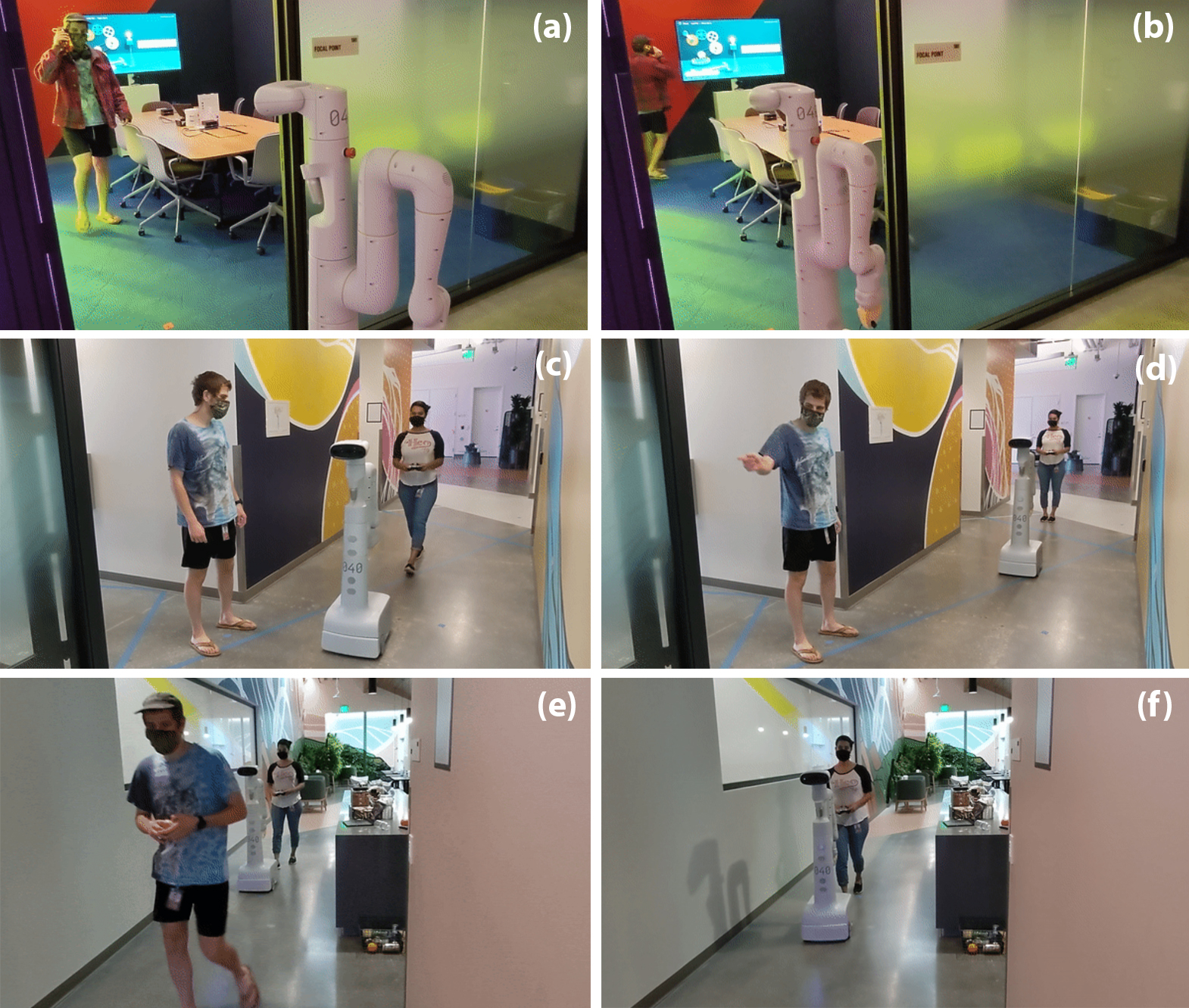} 
  \end{center}
  \vspace{-5mm}
  \caption{Examples of the social scenarios: narrow doorway (a, b), intersection gesture (c, d), and intersection wait (e, f). Each scenario is defined by enacting the social interaction with a human and a robot that is manually controlled by a human operator.}
  \label{fig:social_scenarios_real}   
  \vspace{-6mm}
\end{figure}

\subsection{Social Navigation Scenarios}

For our current benchmark we have selected five social scenarios of human-robot interactions as detailed below. For most of our scenarios we focus on observant and passive robot behavior; i.e. the robot is expected to yield and make room for the human. Consequently, our ideal social navigation policy would generate robot behavior in a way that the human would almost not notice the robot. An illustration of the social scenarios of our benchmark is shown in Fig.~\ref{fig:social_scenarios}. Examples of our real setup for each social scenario are shown in Fig.~\ref{fig:teaser}, ~\ref{fig:social_scenarios_real}, and ~\ref{fig:blind_corner}. 

\textbf{Frontal:} Robot and human are approaching each other from two ends of a straight trajectory; enough space is provided for the robot to yield. Robot and human are walking toward each other and the robot is expected to yield early on to avoid socially-intimidating behavior. Human and robot are alternating their start and end positions. 

\textbf{Intersection Wait}: Robot and human approach each other on perpendicular trajectories. The human does not stop walking down their path. The robot is expected to drive slowly when it approaches the human and it has to come to a complete stop to yield to the human. After the human is out of sight, the robot continues on its trajectory.  

\textbf{Intersection Gesture:} Robot and human approach each other on perpendicular trajectories. Human and robot come to a complete stop. The human recognizes the robot and then gestures -- with a waving hand motion along the trajectory of the robot -- that they yield to the robot. The robot interprets the gesture and continues its path. 

\textbf{Narrow Doorway:} Human and robot cross each other's paths by moving through a narrow doorway. Robot and human alternately start inside or outside a room and try to get in or out. In this scenario, the human or the robot has to yield to the respective other agent. If the robot arrives at the door before the human it is allowed to continue on its path. If the human arrives at the door first, the robot has to wait outside the door and yield to the human. 

\textbf{Blind Corner:} Human and robot cross each other’s paths at a blind corner. Human and robot move down a hallway toward the corner and ‘surprise each other’ by meeting at the corner at the same time. Both agents have to come to a complete stop to then resolve the situation. The robot is expected to either yield to the human after the collision or to avoid the collision by anticipating the situation.

\section{In-Situ Validation}

To validate the social-compliance of a policy we define an in-situ metric based on a questionnaire for each social scenario. We use a five-level Likert scale \cite{likert1932technique} to define 4 - 5 questions for each scenario and ask participants to rate their agreement toward these questions based on the following scale: 1=strongly disagree, 2=disagree, 3=neutral, 4=agree, 5=strongly agree. Additionally, we allow for the rating 0=cannot tell/something went wrong. The questions for each scenario are provided in Table~\ref{tab:questionnaire}: ``How much do you agree with the statement ... ?''. We run the scenario with human and robot and immediately ask the participant to provide the rating before performing another scenario. This allows us to get reliable in-experience ratings.  

\begin{table} [t]
\small
\centering
\scalebox{0.95}{
  \begin{tabular}{l|l}
  \hline
  \multicolumn{2}{l}{\textbf{Frontal Approach}} \\
  \hline
  1 & The robot moved to avoid me.\\
  2* & The robot obstructed my path.  \\
  3 & The robot maintained a safe and comfortable distance \\ 
  &  at all times.\\
  4* & The robot nearly collided with me. \\
  5 & It was clear what the robot wanted to do.\\
  \hline
  \multicolumn{2}{l}{\textbf{Intersection Wait}} \\
  \hline
  6 & The robot let me cross the intersection by maintaining \\ 
  &  a safe and comfortable distance.\\
  7 & The robot changed course to let me pass. \\
  8 & The robot paid attention to what I was doing.\\ 
  9 & The robot slowed down and stopped to let me pass.\\ 
  \hline
  \multicolumn{2}{l}{\textbf{Intersection Gesture}} \\
  \hline  
  10 & The robot maintained a safe and comfortable distance \\
  & at all times. \\
  11 & The robot slowed down and stopped. \\
  12 & The robot followed my command. \\
  13 & I felt the robot paid attention to what I was doing. \\
  \hline
  \multicolumn{2}{l}{\textbf{Narrow Doorway}} \\
  \hline  
  14* & The robot got in my way. \\
  15 & The robot moved to avoid me. \\
  16 & The robot made room for me to enter or exit. \\
  17* & It was clear what the robot wanted to do. \\
  \hline
  \multicolumn{2}{l}{\textbf{Blind Corner}} \\
  \hline    
  18 & The robot moved to avoid me. \\
  19 & The robot stopped to let me pass. \\
  20* & I had to move around the robot. \\
  21* & The robot nearly collided with me head-on. \\
  \hline   
  \end{tabular}
}
\vspace{-1mm}
\caption{\small Questions for each social scenario.}
\vspace{-11mm}
\label{tab:questionnaire}
\end{table}

%% file: sources/05_experiments.tex
\begin{table*}[t]
\small
\centering
\scalebox{0.9}{
  \begin{tabular}{c|c|c|c|c|c|c|c|c|c|c|c|c|c|c|c|c|c|c|c|c|c}
    & \multicolumn{5}{c|}{\textbf{Frontal Approach}} & \multicolumn{4}{c|}{\textbf{Intersection Wait}} & \multicolumn{4}{c|}{\textbf{Intersection Gesture}} & \multicolumn{4}{c|}{\textbf{Narrow Doorway}} & \multicolumn{4}{c}{\textbf{Blind Corner}}\\
   Question   & 1  & 2 & 3 & 4 & 5 & 6 & 7 & 8 & 9 & 10 & 11 & 12 & 13 & 14 & 15 & 16 & 17 & 18 & 19 & 20 & 21\\
   \hline
   \hline
    Location & \multicolumn{5}{c|}{F1} & \multicolumn{4}{c|}{W1} & \multicolumn{4}{c|}{G1} & \multicolumn{4}{c|}{N1} & \multicolumn{4}{c}{B1} \\
   \hline   
   QAVG & 3.1 & 2.6 & 3.2 & 2.7 & 2.7  & 2.2 & 2.4 & 2.4 & 1.3   & 2.1 & 3.8 & 2.2 & 2.1  & 4.1 & 2.2 & 2.0 & 4.4   & 3.4 & 2.0 & 2.9 & 2.0 \\
   \hline
   SAVG & \multicolumn{5}{c|}{2.9} & \multicolumn{4}{c|}{2.1} & \multicolumn{4}{c|}{2.6} & \multicolumn{4}{c|}{1.9} & \multicolumn{4}{c}{3.1} \\
   \hline
   STD & \multicolumn{5}{c|}{1.7} & \multicolumn{4}{c|}{1.4} & \multicolumn{4}{c|}{0.6} & \multicolumn{4}{c|}{1.4} & \multicolumn{4}{c}{1.5} \\   
  \hline
  \hline
 Location & \multicolumn{5}{c|}{F2} & \multicolumn{4}{c|}{W2} & \multicolumn{4}{c|}{G2} & \multicolumn{4}{c|}{N2} & \multicolumn{4}{c}{B2} \\
   \hline   
   QAVG & 2.7 & 3.6 & 3.3 & 2.7 & 3.0  & 2.5 & 2.8 & 2.2 & 1.2   & 3.2 & 1.8 & 1.4 & 2.0  & 3.7 & 1.6 & 1.3 & 4.3   & 1.9 & 2.6 & 3.2 & 2.7 \\
   \hline
   SAVG & \multicolumn{5}{c|}{2.9} & \multicolumn{4}{c|}{1.3} & \multicolumn{4}{c|}{2.1} & \multicolumn{4}{c|}{1.7} & \multicolumn{4}{c}{2.7} \\
   \hline
   STD & \multicolumn{5}{c|}{1.7} & \multicolumn{4}{c|}{1.3} & \multicolumn{4}{c|}{1.2} & \multicolumn{4}{c|}{1.2} & \multicolumn{4}{c}{1.5} \\   
  \end{tabular}
}
\vspace{-1mm}
\caption{\small Five social scenarios each defined for two locations.}
\vspace{-7mm}
\label{tab:results_all}
\end{table*}

\begin{table}[t]
\small
\centering
\scalebox{0.6}{
  \begin{tabular}{l|c|c|c|c|c|c|c|c|c|c|c|c|c|c|c}
    & \multicolumn{5}{c|}{\textbf{Participant 1}} & \multicolumn{5}{c|}{\textbf{Participant 2}} & \multicolumn{5}{c}{\textbf{Participant 3}} \\
    Question   & 1  & 2 & 3 & 4 & 5 & 1 & 2 & 3 & 4 & 5   & 1 & 2 & 3 & 4 & 5 \\
   \hline
   \hline
   QAVG & 2.8 & 3.8 & 2.0 & 2.8 & 3.4  
        & 2.6 & 3.3 & 2.0 & 3.5 & 2.2   
        & 2.6 & 2.9 & 2.6 & 2.9 & 2.5 \\
   \hline
   SAVG & \multicolumn{5}{c|}{2.5} & \multicolumn{5}{c|}{2.7} & \multicolumn{5}{c}{2.8} \\
   \hline
   STD & \multicolumn{5}{c|}{1.3} & \multicolumn{5}{c|}{1.1} & \multicolumn{5}{c}{1.1} \\
  \hline
  \end{tabular}
}
\caption{\small  Frontal Approach: Three Different Participants.}
\label{tab:results_subject}

\small
\centering
\scalebox{0.6}{
  \begin{tabular}{l|c|c|c|c|c|c|c|c|c|c|c|c|c|c|c}
    & \multicolumn{5}{c|}{\textbf{Set 1}} & \multicolumn{5}{c|}{\textbf{Set 2}} & \multicolumn{5}{c}{\textbf{Set 3}} \\
    Question   & 1  & 2 & 3 & 4 & 5 & 1 & 2 & 3 & 4 & 5   & 1 & 2 & 3 & 4 & 5 \\
   \hline
   \hline
   QAVG & 2.9 & 2.2 & 2.0 & 2.0 & 3.4  
        & 2.6 & 3.1 & 2.6 & 1.0 & 2.5   
        & 2.3 & 3.0 & 2.4 & 3.0 & 2.3 \\
   \hline
   SAVG & \multicolumn{5}{c|}{2.5} & \multicolumn{5}{c|}{2.4} & \multicolumn{5}{c}{2.6} \\
   \hline
   STD & \multicolumn{5}{c|}{1.3} & \multicolumn{5}{c|}{1.1} & \multicolumn{5}{c}{1.0} \\
  \hline
  \end{tabular}
}
\caption{\small  Frontal Approach: Three Sets.}

\label{tab:results_sets}

\small
\centering
\scalebox{0.85}{
  \begin{tabular}{l|c|c|c|c|c|c|c|c|c|c}
    & \multicolumn{5}{c|}{\textbf{Policy 1 (MPC)}} & \multicolumn{5}{c}{\textbf{Policy 2 (MPC + BC)}} \\
    Question   & 1  & 2 & 3 & 4 & 5 & 1 & 2 & 3 & 4 & 5   \\
  \hline
  \hline
  QAVG & 2.3 & 4.3 & 2.0 & 3.9 & 2.6  
        & 4.6 & 1.7 & 4.3 & 1.6 & 4.3 \\
  \hline
  SAVG & \multicolumn{5}{c|}{2.2} & \multicolumn{5}{c}{4.3} \\
  \hline
  STD & \multicolumn{5}{c|}{1.1} & \multicolumn{5}{c}{1.2}  \\
  \hline
  \end{tabular}
}
\caption{\small  Frontal Approach: Two Policies.}
\label{tab:results_policy}
\vspace{-10mm}
\end{table}

\section{Experiments}

To begin validating our benchmark, we have conducted experiments demonstrating that our setup is scalable, reliable and repeatable. For most of the  experiments we use a simple iLQR-based model predictive controller (MPC) to generate linear and angular velocity commands for our robot (provided by Everyday Robots, Alphabet). We repeatedly run the policy and ask the human participants to answer the questionnaire (Table~\ref{tab:questionnaire}) after each run.  

In Table~\ref{tab:results_all} we show the results of running the MPC policy against all of our five social scenarios in both of the defined locations (Fig.~\ref{fig:floor_layout}).  For this experiment, we recorded 10 runs for each social scenario in both of the defined locations. We report the per question average (QAVG), as well as the overall average (SAVG) along with the standard deviation for each scenario. For negatively formulated questions (labeled with a `*' in  Table~\ref{tab:questionnaire}) we reverse coded the ratings to make them comparable to the positively formulated ones. Across the different scenarios we obtain similar average ratings. This suggests that our questionnaire based metric can be used to obtain meaningful results for different social scenarios. Moreover, we observed that the measurements for the same social scenario in different locations corresponded with each other. This indicates that our setup can be replicated for the same social scenarios in other locations (e.g. other labs) or for other social scenarios. One run of a social scenario commonly takes 15-45 seconds, while answering the 4-5 questions for each scenario requires 20-40 seconds.

To test our setup for individual human biases, we ran the same Location~(F2) frontal approach scenario 30 times with three different participants. 
Table~\ref{tab:results_subject} shows that for each participant we obtained similar average ratings for each question and for the entire social scenario.
This suggests that our questionnaire-based metric provides good inter-rater reliability.

To measure the variance of our setup when running the same experiment at different dates and times, we ran the Location~(F2) frontal approach scenario 90 times with the same human participant. Table~\ref{tab:results_sets} shows similar average ratings and standard deviations across the different runs of the same social scenario. 
This suggests that our questionnaire based metric can be used to obtain reliable measurements across different validation runs. 
Furthermore, this suggests that it may not be necessary to capture large quantities of runs to obtain reliable results but that batches of up to 30 runs already provide meaningful results. 

To test our benchmark for different polices we compare the MPC policy with a behavioral cloning (BC) policy. Here we captured 300 trajectories of expert data of the frontal approach scenario and trained a convolutional neural network to predict intermediate waypoints for the MPC policy. We use this policy to generate socially-compliant behavior.  Table~\ref{tab:results_policy} shows the average questionnaire ratings of 20 runs of each policy. These results indicate that we are able to measure the capabilities of this more advanced policy compared to the common MPC policy (indicated by the higher averages). 

\begin{figure}[t]   
  \begin{center}
  \includegraphics[width=1.0\linewidth]{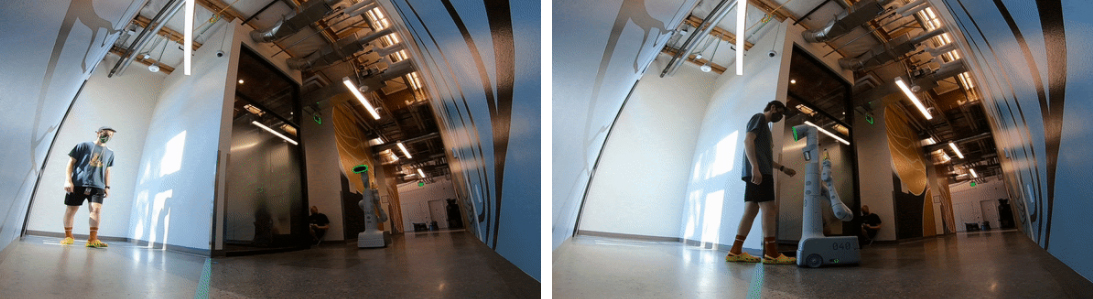} 
  \end{center}
  \vspace{-5mm}
  \caption{Validation run of the blind corner scenario. Robot and human are moving in opposite directions around a blind corner on a colliding path. The MPC policy is not able to anticipate the human interaction causing a collision -- the robot drove onto the person's foot. Consequently the rating for this social interaction was: Q18=strongly disagree, Q19=disagree, Q20=neutral, Q21=strongly agree.}
  \label{fig:blind_corner}   
  \vspace{-6mm}
\end{figure}

In Fig.~\ref{fig:blind_corner} we show a validation run of our blind corner scenario that highlights the advantages of obtaining in-situ ratings from human participants. For this run, the robot briefly collided with the human, touching the person's foot. While this event is hardly noticeable for an external observer or in an ex-situ setting (Fig.~\ref{fig:blind_corner}, right) it generated a clear negative response when the participant provided their ratings.  

\vspace{-3mm}

%% file: sources/06_conclusion.tex
\vspace{2mm}
\section{Conclusion}
\vspace{-1mm}
We have introduced a novel protocol for establishing a benchmark for social navigation scenarios. The proposed approach is based on defining a set of common social interactions that occur for navigation tasks. Each social scenario is defined in a canonical manner to support the repeated validation of policies. Additionally, we have proposed and piloted a questionnaire-based metric to obtain in-situ ratings of human participants that allow us to assess the social compliance of navigation policies. We only rely on a lightweight specification for each social scenario. Therefore, our benchmark and our questionnaire can readily be extended by additional scenarios. As future work, we plan to extend our benchmark and to use it to validate existing and novel socially-compliant navigation policies. 
\vspace{-1mm}
\section{Acknowledgments}
\vspace{-2mm}
We thank our robot operators April Zitkovich, Jake Lee, Khem Holden, Rosario Jauregui Ruano, and Diego Reyes for their diligent work collecting all social navigation data samples reported in the paper.